\RequirePackage{fix-cm}

\documentclass[twocolumn]{svjour3}         
\smartqed  

\usepackage{ifpdf}
\usepackage{adjustbox}
\usepackage{dsfont}
\usepackage[frozencache=true,cachedir=minted-cache]{minted}
\usepackage{multirow}
\usepackage{subcaption}
\usepackage{url}
\usepackage{xspace}

\usepackage{tikz}
\usetikzlibrary{calc,trees,positioning,arrows,chains,shapes.geometric,%
    decorations.pathreplacing,decorations.pathmorphing,shapes,%
    matrix,shapes.symbols}

\tikzset{>=stealth',
  io/.style={
    minimum height=3em,
    text centered,
    on chain},
  bstep/.style={
    rectangle,
    rounded corners,
    draw=black, dashed, 
    text width=7em,
    minimum height=3em,
    text centered,
    on chain},
  block/.style={
    rectangle,
    rounded corners,
    draw=black, 
    text width=7em,
    minimum height=3em,
    text centered,
    on chain}}

\newcommand*{\affaddr}[1]{#1}
\newcommand*{\affmark}[1][*]{\textsuperscript{#1}}

\newcommand{\cFeatures}{\texttt{Features}\xspace}
\newcommand{\cFeaturesCollection}{\texttt{FeaturesCollection}\xspace}
\newcommand{\cProcessor}{\texttt{Processor}\xspace}

\newcommand{\cAudio}{\texttt{Audio}\xspace}
\newcommand{\cUtterances}{\texttt{Utterances}\xspace}
\newcommand{\true}{\texttt{True}}
\newcommand{\false}{\texttt{False}}

\begin{document}

\title{Shennong: a Python toolbox for audio speech features extraction}


\author{
    Mathieu Bernard\protect\affmark[*]\affmark[1] \and
    Maxime Poli\protect\affmark[*]\affmark[1] \and
    Julien Karadayi\protect\affmark[1] \and
    Emmanuel Dupoux\protect\affmark[1,2] 
    }

\authorrunning{Bernard et al.}

\institute{
    Mathieu Bernard -- Inria, 2 rue Simone Iff, 75012 Paris, France.\\
    \email{mathieu.a.bernard@inria.fr}\\
    \affaddr{\affmark[*]Equal contribution as first authors.}\\
    \affaddr{\affmark[1]EHESS, ENS, PSL Research University, CNRS and Inria.}\\
    \affaddr{\affmark[2]Facebook AI Research.}
    }

\date{Received: date / Accepted: date}

\maketitle

\begin{abstract}
We introduce Shennong, a Python toolbox and command-line utility for speech features extraction. It implements a wide range of well-established state of art algorithms including spectro-temporal filters such as Mel-Frequency Cepstral Filterbanks or Predictive Linear Filters, pre-trained neural networks, pitch estimators as well as speaker normalization methods and post-processing algorithms. Shennong is an open source, easy-to-use, reliable and extensible framework. The use of Python makes the integration to others speech modeling and machine learning tools easy. It aims to replace or complement several heterogeneous software, such as Kaldi or Praat. After describing the Shennong software architecture, its core components and implemented algorithms, this paper illustrates its use on three applications: a comparison of speech features performances on a phones discrimination task, an analysis of a Vocal Tract Length Normalization model as a function of the speech duration used for training and a comparison of pitch estimation algorithms under various noise conditions.
\end{abstract}

\section{Introduction}
\label{intro}

Automatic processing of speech is at the heart of a wide range of applications: speech to text \cite{benzeghiba2007automatic}, speaker identification \cite{tirumala2017speaker}, emotion recognition \cite{koolagudi2012emotion} or speaker diarization \cite{ryant2019dihard,ryant2020dihard}. It is also applied to a variety of contexts such as multilingual models \cite{bottleneck,bottleneck2}, low-resource languages \cite{dunbar2017zero,dunbar2020zero}, pathological speech \cite{orozco2016automatic,riad2020vocal} or, more recently, end-to-end deep learning models \cite{saeed2020contrastive,zeghidour2018endtoend}.
All of those applications rely on some representation or \emph{features} of the speech signal, \emph{i.e.} a transformation of the raw audio signal which carries informative and/or discriminative information, usually in the time-frequency domain, that can further be processed and analyzed. As so, features extraction is the first step of most speech processing pipelines.

A lot of speech features extraction softwares have been proposed over time, with a huge number of implementations in different programming languages. Among them, some tools gained a wide audience. Kaldi \cite{kaldi} is an Automatic Speech Recognition toolkit that covers every aspect of this topic, from language modeling to decoding and features extraction. Written in C++, it supports a collection of state of the art recipes as Bash scripts. Whereas it is very reliable and efficient, it is hard to use and to embed in third party tools for non-technical users. Praat \cite{Praat} is another popular software used for speech analysis in phonetics, particularly for speech annotation. It can be used from a graphical user interface or from a custom scripting language. It includes basic spectro-temporal analysis, such as spectrogram, cochleogram and pitch analysis. OpenSMILE \cite{opensmile} is another features extraction package. Designed for real time processing, it focuses on audio signal but is generic enough to be used for visual or physiological signals as well. Usable from command-line of from wrapper in various programming languages, it's generic approach make it hard to use and configure. Finally, Surfboard \cite{surfboard} is a Python toolbox dedicated to speech features extraction. It is oriented toward medical applications and implements numbers of specialized markers. Both OpenSMILE and Surfboard are very suitable tools but they lack of general purpose features such as speaker normalization, and they do not propose fine grained parameters as offered by Kaldi.

The main objective of the Shennong toolbox is to provide reference implementations of speech features extraction algorithms in an easy-to-use and reliable framework. The use of Python makes it easy to integrate Shennong with modern machine learning tools such  as scikit-learn \cite{scikit-learn}, Pytorch \cite{pytorch} and Tensorflow \cite{tensorflow}. Another design goal for Shennong is that it can be used both by casual users, with provided pre-configured pipelines, and power users, being entirely customizable and easily extensible. Finally, by distributing such a tool to the community, our objective is also to reduce the use of heterogeneous features extraction implementations used in the literature, and therefore improving replicability and comparability of studies.

This paper is structured as follows. Section \ref{sec:toolbox} describes the speech processing algorithms available in Shennong and the architecture of the toolbox, from low-level components to high-level user interfaces. It also introduces simple usage examples. Section \ref{sec:experimentation} exposes three applications. First, the features extraction algorithms implemented in Shennong are compared on a phoneme discrimination task. An analysis of Vocal Track Length Normalization models quality as a function of speech duration used for training is then proposed. The final experiment compares three pitch estimation algorithms under different noisy conditions.

\section{The Shennong toolbox}
\label{sec:toolbox}

Shennong\footnote{Shennong is named after the so-called Chinese Emperor that popularized the tea according to Chinese Mythology. This is a reference to Kaldi, a speech recognition toolkit on which Shennong is built, and a legendary Ethiopian goatherd who discovered the coffee plant.} is a Python toolbox for easy-to-use, reliable and reproducible speech features extraction. The package is distributed as open-source software\footnote{\url{https://github.com/bootphon/shennong}} under a GPL3 licence.
It is available for Linux and MacOS systems, as well as a Docker image \cite{docker}, which can be deployed on Windows.  It can be used as a Python library and being integrated in third-party applications, or used directly from the command line and called from bash scripts. The code follows high quality standards in terms of software development, testing and documentation. It's modular design makes it easily extensible. The continuous development of Shennong leads to successive versions being released. This paper is based on the version 1.0.

\subsection{Implemented algorithms}
\label{sec:models}

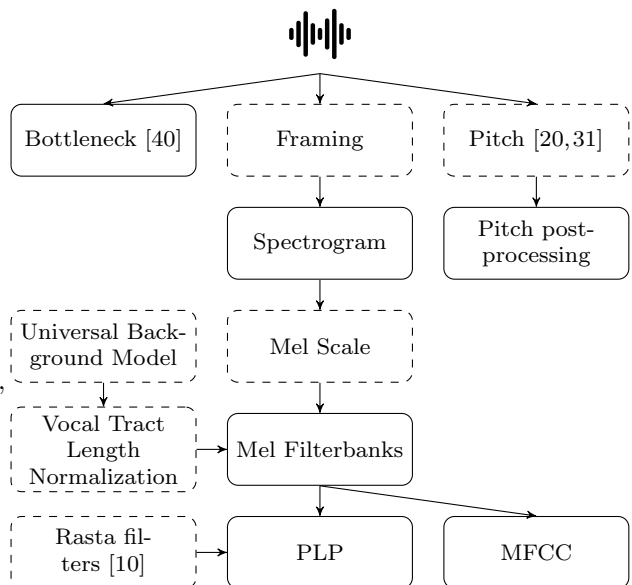
\begin{figure}[t]
\begin{center}
\begin{tikzpicture}[node distance=.4cm, start chain=going below]
  \node[io] (speech) {\includegraphics[scale=0.05]{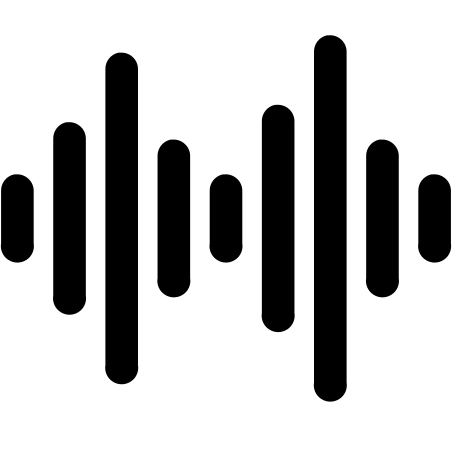}};
  \node[bstep] (frame) {Framing};
  \node[block, below=.4cm of frame] (spec) {Spectrogram};
  \node[bstep, below=.4cm of spec] (mel) {Mel Scale};
  \node[block] (fbank) {Mel Filterbanks};
  \node[block, below=.4cm of fbank] (plp) {PLP};
  \node[bstep, left=.4cm of plp] (rasta) {Rasta filters \cite{rastamat}};
  \node[block, right=.4cm of plp] (mfcc) {MFCC};
  \node[block, left=.4cm of frame] (bottle) {Bottleneck \cite{bottleneck2}};
  \node[bstep, right=.4cm of frame] (pitch) {Pitch \cite{crepe,kaldi}};
  \node[block, below=.4cm of pitch] (postpitch) {Pitch post-processing};
  \node[bstep, left=.4cm of mel] (ubm) {Universal Background Model};
  \node[bstep, left=.4cm of fbank] (vtln) {Vocal Tract Length Normalization};
  \draw[->] (speech.south) -- (frame.north);
  \draw[->] (frame.south) -- (spec.north);
  \draw[->] (spec.south) -- (mel.north);
  \draw[->] (mel.south) -- (fbank.north);
  \draw[->] (ubm.south) -- (vtln.north);
  \draw[->] (vtln.east) -- (fbank.west);
  \draw[->] (fbank.south) -- (mfcc.north);
  \draw[->] (fbank.south) -- (plp.north);
  \draw[->] (rasta.east) -- (plp.west);
  \draw[->] (speech.south) -- (bottle.north);
  \draw[->] (speech.south) -- (pitch.north);
  \draw[->] (pitch.south) -- (postpitch.north);
\end{tikzpicture}
\end{center}
\caption{\label{fig:processors}Features extraction algorithms hierarchy, from speech signal to usable features (blocks in full lines) with intermediate or optional steps (blocks in dashed lines). When no reference is present, the blocks are implemented after Kaldi \cite{kaldi}. The available parameters for each block are detailed in Table \ref{tab:parameters}. Post-processing algorithms, such as Delta and Cepstral Mean Variance Normalization, are not represented in this diagram (see text for details).}
\end{figure}

\begin{table*}[!ht]
    \centering
    \begin{adjustbox}{totalheight=\textheight-4\baselineskip}
    \begin{tabular}{|c|c|c|l|}
        \hline
        \textbf{Algorithm} & \textbf{Parameter} & \textbf{Default} & \textbf{Comment} \\
        \hline
        Bottleneck & \texttt{weights} & \texttt{BabelMulti} & Pretrained network to use, in \texttt{FisherMono}, \texttt{FisherMulti} or \texttt{BabelMulti} \\
        & \texttt{dither} & 0.1 & Amount of dithering to add \\
        \hline
        Framing & \texttt{sample\_rate} & 16000 & Sampling frequency in Hz\\
        & \texttt{frame\_shift} & 0.01 & Frame shift in second \\
        & \texttt{frame\_length} & 0.025 & Frame length in second \\
        & \texttt{dither} & 0.1 & Amount of dithering to add \\
        & \texttt{preemph\_coeff} & 0.97 & Signal preemphasis coefficient\\
        & \texttt{remove\_dc\_offset} & \true & Whether to subtract mean on each frame \\
        & \texttt{window\_type} & \texttt{povey} & Window to use in \texttt{hamming}, \texttt{hanning}, \texttt{povey}, \texttt{rectangular} or \texttt{blackman} \\
        & \texttt{snip\_edges} & \true & If true, output only frames that completely fit in the input signal\\
        \hline
        Spectrogram & \multicolumn{1}{l|}{\emph{all from Framing plus...}}&&\\
        & \texttt{energy\_floor} & 0.0 & Absolute floor on energy\\
        & \texttt{raw\_energy} & \true & When true, compute energy before preemphasis and windowing\\
        \hline
        Mel Scale & \multicolumn{1}{l|}{\emph{all from Framing plus...}}&&\\
        & \texttt{num\_bins} & 23 & Number of triangular mel-frequency bins\\
        & \texttt{low\_freq} & 20 & Low cutoff frequency for mel bins in Hz\\
        & \texttt{high\_freq} & 0 & High cutoff frequency for mel bins in Hz\\
        & \texttt{vtln\_low} & 100 & Low inflection point in VTLN in Hz\\
        & \texttt{vtln\_high} & -500 & High inflection point in VTLN in Hz\\
        \hline
        Filterbank & \multicolumn{1}{l|}{\emph{all from Mel Scale plus...}}&&\\
        & \texttt{use\_energy} & False & Add an extra dimension with energy to the filterbank output\\
        & \texttt{energy\_floor} & 0.0 & Absolute floor on energy\\
        & \texttt{raw\_energy} & \true & When true, compute energy before preemphasis and windowing\\
        & \texttt{use\_log\_fbank} & \true & Whether to produce log or linear filterbanks\\
        & \texttt{use\_power} & \true & Whether to use power or magnitude\\
        \hline
        MFCC & \multicolumn{1}{l|}{\emph{all from Mel Scale plus...}}&&\\
        & \texttt{num\_ceps} & 13 & Number of cepstra, including C0\\
        & \texttt{use\_energy} & False & Add an extra dimension with energy to the filterbank output\\
        & \texttt{energy\_floor} & 0.0 & Absolute floor on energy\\
        & \texttt{raw\_energy} & \true & When true, compute energy before preemphasis and windowing\\
        & \texttt{cepstral\_lifter} & 22.0 & Constant that controls scaling of MFCCs\\
        \hline
        PLP & \multicolumn{1}{l|}{\emph{all from Mel Scale plus...}}&&\\
        & \texttt{rasta} & \false & Whether to do RASTA filtering\\
        & \texttt{lpc\_order} & 12 & Order of LPC analysis\\
        & \texttt{num\_ceps} & 13 & Number of cepstra, including C0\\
        & \texttt{use\_energy} & False & Add an extra dimension with energy to the filterbank output\\
        & \texttt{energy\_floor} & 0.0 & Absolute floor on energy\\
        & \texttt{raw\_energy} & \true & When true, compute energy before preemphasis and windowing\\
        & \texttt{compress\_factor} & 1/3 & Compression factor\\
        & \texttt{cepstral\_lifter} & 22.0 & Constant that controls scaling of PLPs\\
        & \texttt{cepstral\_scale} & 1.0 & Cepstral constant in PLP computation\\
        \hline
        Pitch& \texttt{sample\_rate} & 16000 & Sampling frequency in Hz\\
        (Kaldi algorithm)& \texttt{frame\_shift} & 0.01 & Frame shift in second \\
        & \texttt{frame\_length} & 0.025 & Frame length in second \\
        & \texttt{min\_f0} & 50 & Minimum F0 to search for in Hz\\
        & \texttt{max\_f0} & 400 & Maximum F0 to search for in Hz\\
        & \texttt{soft\_min\_f0} & 10 & Minimum F0 to search for in Hz, applied in soft way\\
        & \texttt{penalty\_factor} & 0.1 & Cost factor for F0 change\\
        & \texttt{lowpass\_cutoff} & 1000 & Cutoff frequency for low-pass filter in Hz \\
        & \texttt{resample\_freq} & 4000 & Downsampling frequency in Hz\\
        & \texttt{delta\_pitch} & 0.005 & Smallest relative change in pitch that the algorithm measures\\
        & \texttt{nccf\_ballast} & 7000 & Increasing this factor ensure pitch continuity in unvoiced regions\\
        \hline
        Pitch& \texttt{model\_capacity}& \texttt{full}& Pretrained model to use, in \texttt{tiny}, \texttt{small}, \texttt{medium}, \texttt{large} or \texttt{full}\\
        (CREPE algorithm)& \texttt{frame\_shift} & 0.01 & Frame shift in second \\
        & \texttt{frame\_length} & 0.025 & Frame length in second \\
        & \texttt{viterbi} & \true & Whether to apply viterbi smoothing to the estimated pitch curve\\
        & \texttt{center} & \true & Whether to center the window on the current frame \\
        \hline
        Pitch & \texttt{pitch\_scale} & 2.0 & Scaling factor for the final normalized log-pitch value\\
        (post-processing)& \texttt{pov\_scale} & 2.0 & Scaling factor for final probability of voicing feature\\
        & \texttt{delta\_pitch\_scale} & 10.0 & Term to scale the final delta log-pitch feature\\
        & \texttt{delta\_pitch\_noise\_stddev} & 0.005 & Standard deviation for noise we add to the delta log-pitch\\
        & \texttt{delta\_window} & 2 & Number of frames on each side of central frame\\
        & \texttt{delay} & 0 & Number of frames by which the pitch information is delayed\\
        \hline
        Universal & \texttt{num\_gauss} & 64 & Number of Gaussians in the model\\
        Background Model& \texttt{num\_iters} & 4 & Number of training iterations\\
        & \texttt{initial\_gauss\_proportion} & 0.5 & Proportion of Gaussians to start with in initialization phase\\
        & \texttt{num\_iters\_init} & 20 & Number of E-M iterations for model initialization \\
        & \texttt{num\_frames} & $5.10^5$ & Maximum num-frames to keep in memory for model initialization \\
        & \texttt{min\_gaussian\_weight} & $10^{-4}$ & Minimum weight below which a Gaussian is not updated\\
        & \texttt{remove\_low\_count\_gaussians} & \false & Remove Gaussians with a weight below \texttt{min\_gaussian\_weight}\\
        \hline
        Vocal Tract Length& \multicolumn{1}{l|}{\emph{all from UBM plus...}}&&\\
        Normalization& \texttt{num\_iters} & 15 & Number of training iterations\\
        & \texttt{min\_warp} & 0.85 & Minimum warp considered\\
        & \texttt{max\_warp} & 1.15 & Maximum warp considered\\
        & \texttt{warp\_step} & 0.01 & Warp step\\
        & \texttt{logdet\_scale} & 0.0 & Scale on log-determinant term in auxiliary function\\
        & \texttt{norm\_type} & \texttt{offset} & Type of fMMLR applied, in \texttt{offset}, \texttt{none} or \texttt{diag}\\
        \hline
    \end{tabular}
    \end{adjustbox}
    \caption{Parameters of the features extraction algorithms implemented in Shennong. Zero or negative frequencies are relative to the Nyquist frequency.}
    \label{tab:parameters}
\end{table*}

The algorithms available in Shennong are presented in Figure \ref{fig:processors}. Most of them are implemented after Kaldi \cite{kaldi}, using the pykaldi Python wrapper \cite{pykaldi}. The algorithms are provided with all the parameters given by original implementations and default values suitable for most cases, as detailed in Table \ref{tab:parameters}. All the implemented algorithms have been carefully and extensively tested so as to replicate original implementations. The remaining of this section introduces those algorithms, for which a comparison is provided in  Section \ref{sec:features}.

Spectro-temporal representations are widely used methods based on short-term spectral analysis. In Shennong they include Spectrogram, Mel-Filterbanks, Mel-Frequency Cepstrum Coefficients (MFCC) and Perceptual Linear Predictive filters (PLP). Starting from raw speech, the signal is first split into overlapping frames, on which the power spectrum is computed. The power spectrum, along with signal energy, and optionally expressed in the log domain, is used to generate the Spectrogram features. The Mel Filterbanks are then obtained by applying a mel scale to the power spectrum. Finally MFCC and PLP are obtained with further processing in the cepstral domain.
Rasta filters are optionnal bandpass filters that can be applied to PLP features \cite{hermansky1990perceptual,hermansky1994rasta,hermansky1991rasta}, so as to make them more robust to linear spectral distortions due to the communication channel.

Vocal Track Length Normalization (VTLN) \cite{kim2004vtln,povey2010vtln} is a normalization technique used to reduce inter-speaker variability. It can be applied to Mel based representations, namely Mel Filterbanks, MFCC and PLP features. It consists of a model based estimation of speaker-specific linear transforms of the power spectrum that scale the mel filters center frequencies and bandwidths. It requires an Universal Background Model to be trained in order to estimates a VTLN warp coefficient per speaker, that is applied to features for normalization. The training is unsupervised and does not requires any annotation or phonetic transcription. The effectiveness of VTLN is demonstrated on Section \ref{sec:features} and a study on the amount of data required to train a VTLN model is provided in Section \ref{sec:vtln}.

The Bottleneck features \cite{bottleneck,bottleneck2} relies on convolutional neural networks pre-trained for phones recognition. Three networks are available: monophone and triphone states, both trained on US English from the Fisher dataset \cite{cieri2004fisher}, and a multilingual triphone states network trained on 17 languages from the Babel dataset \cite{harper2013babel}.

Shennong also implements two algorithms for pitch estimation. The first one from Kaldi \cite{pitchkaldi} is based on normalized a cross-correlation of the input signal and outputs for each frame a pitch estimate along with a probability of voicing. The second algorithm is Convolutional REpresentation for Pitch Estimation (CREPE) \cite{crepe} and is based on a convolutional neural network pre-trained on music datasets \cite{crepe,pyin}. The CREPE algorithm is made fully compatible with the Kaldi one by turning the maximum of the network activation matrix into a probability of voicing and by interpolating pitch for frames with low confidence. Finally, a post-processing step, common to both algorithms, normalizes the pitch estimates, convert them to log domain and extract their first-order derivative. A comparison of those algorithms is proposed Section \ref{sec:pitch}.

Finally Shennong also provides post-processors that normalize or add information on extracted features. Delta computes the $n^{th}$ order derivative of any raw features. Voice Activity Detection (VAD) is a simple energy based method that makes binary decisions, without any notion of continuity, that can be used to filter out silences. Cepstral Mean Variance Normalization (CMVN) normalizes features to a zero mean and unitary variance. It can be applied on a per-frame, per-utterance or per-speaker basis.

\subsection{Low-level software architecture}
\label{sec:architecture}

\begin{table}[t]
    \centering
    \begin{tabular}{|c|c|c|c|}
        \hline
        \textbf{Format} & \textbf{File size} & \textbf{Write time} & \textbf{Read time} \\
        \hline
        pickle & 883 MB & 0:00:07 & 0:00:05 \\
        h5features & 873 MB & 0:00:21 & 0:00:07 \\
        numpy & 869 MB & 0:02:30 & 0:00:22 \\
        matlab & 721 MB & 0:00:59 & 0:00:11 \\
        kaldi & 1.3 GB & 0:00:06 & 0:00:07 \\
        csv & 4.8 GB & 0:03:02 & 0:03:11 \\
        \hline
    \end{tabular}
    \caption{File formats supported by Shennong for reading and writing a \cFeaturesCollection. The read/write times and file size have been obtained on MFCC features computed on the Buckeye English Corpus \cite{buckeye} (40 speakers, about 38 hours of speech in 254 files) using a Linux machine with a Intel Xeon CPU, 16 GB RAM and a SSD hard drive.}
    \label{tab:serializers}
\end{table}

Shennong is built on few low-level components, namely Python classes, that user can use to configure and run a features extraction pipeline.

The \cAudio class is the interface with raw audio data and is the input of all pipelines implemented in Shennong. It is used to load audio files as numpy arrays, resample and manipulate them. It supports multiple audio file formats such as WAV or FLAC. The \cUtterances class provides a high-level view of speech fragments as it handles a collection of \cAudio instances, each one with an attached identifier, speaker information and optional onset and offset times.

The \cFeatures class is the output returned by processing algorithms. It stores three attributes: a data array, a time array and some properties. Data is a numpy array of shape $[m,n]$ with $m$ being the number of frames on the temporal axis and $n$ being the features dimension, usually along the frequency axis. The time array stores the timestamps of each frame either as a single value corresponding to the center time of each frame, with a shape $[m, 1]$, or as a pair of onset/offset times with a shape $[m, 2]$. Several \cFeatures instances sharing the same time values can be concatenated over the frequency axis, so as to obtain composite data within the same array, \emph{e.g.} MFCC and pitch. Finally, the properties record details of the extraction pipeline such as the name of the input audio file and processing parameters values.

The \cFeatures class is designed to store a single matrix corresponding to a single \cAudio. Several \cFeatures are usually grouped into a \cFeaturesCollection, for instance to easily manage a whole dataset represented as a \cUtterances. This class indexes \cFeatures by name and allows to save and load features to/from various file formats (Table \ref{tab:serializers}). The \emph{pickle} format is the native Python one, it is very fast in both writing and reading times and should be the preferred format for little to medium dataset. The \emph{h5features} format \cite{h5features} is specifically designed to handle very large datasets as it allows partial writing and reading of data larger than RAM. The formats \emph{numpy}, \emph{matlab} and \emph{kaldi} propose compatibility layers to those respective tools. Finally the \emph{csv} format stores features into plain text CSV files, one file per \cFeatures in the collection, along with the features properties in JSON format.

The features extraction algorithms are abstracted by the \cProcessor class (see Figure \ref{fig:processors}). Therefore all algorithms implemented in Shennong expose an homogeneous interface to the user: the parameters are specified in the constructor and a \texttt{process()} method takes \cAudio or \cFeatures as input and returns \cFeatures. A generic method \texttt{process\_all()} is also provided to compute features from a whole \cUtterances in a single call, using parallel jobs and returning a \cFeaturesCollection.

\begin{figure*}[t]
    \centering
    \begin{subfigure}{\textwidth}
    \begin{minted}[frame=lines,linenos]{python}
from shennong import Audio, FeaturesCollection
from shennong.processor import MfccProcessor

# load the input WAV file
audio = Audio.load('test.wav')

# extract MFCCs with default parameters
mfcc = MfccProcessor().process(audio)

# save the features as a numpy .npz file
FeaturesCollection({'mfcc': mfcc}).save('mfcc.npz')
    \end{minted}
    \caption{\label{lst:mfcc_simple}MFCC extraction in Python, using the low-level API.}
    \vspace*{1em}
    \end{subfigure}
    \begin{subfigure}{\textwidth}
    \begin{minted}[frame=lines,linenos]{python}
from shennong import pipeline

# generate a pipeline configuration with MFCC and pitch from Kaldi
# (user can then edit parameters in config)
config = pipeline.get_default_config('mfcc', with_pitch='kaldi')

# defines three utterances from two speakers
utterances = [
    ('utterance1', '/path/to/wav1.wav', 'speaker1'),
    ('utterance2', '/path/to/wav2.wav', 'speaker1'),
    ('utterance3', '/path/to/wav3.wav', 'speaker2')]

# apply the configured pipeline on the utterances, run on 3 CPU cores
# and save the extracted features to a numpy format
pipeline.extract_features(config, utterances, njobs=3).save('features.npz')
    \end{minted}
    \caption{\label{lst:pipeline_python}MFCC and pitch extraction in Python, using an extraction pipeline.}
    \vspace*{1em}
    \end{subfigure}

    \begin{subfigure}{\textwidth}
    \begin{minted}[frame=lines,linenos]{bash}
# generate a pipeline configuration with MFCC and pitch from Kaldi
# (user can then edit parameters in config.yaml)
speech-features config mfcc --pitch kaldi -o config.yaml

# defines three utterances from two speakers
echo "utterance1 /path/to/wav1.wav speaker1" > utterances.txt
echo "utterance2 /path/to/wav2.wav speaker1" >> utterances.txt
echo "utterance3 /path/to/wav3.wav speaker2" >> utterances.txt

# apply the configured pipeline on the utterances, run on 3 CPU cores
# and save the extracted features to a numpy format
speech-features extract --njobs 3 config.yaml utterances.txt features.npz
    \end{minted}
    \caption{\label{lst:pipeline_bash}MFCC and pitch extraction from command line, using an extraction pipeline.}
    \end{subfigure}
    \caption{\label{fig:use_example}Examples of use of Shennong. In (\ref{lst:mfcc_simple}) MFCC are extracted and saved from an input audio file. The features have 13 dimensions, which is the default number of Mel coefficients. In (\ref{lst:pipeline_python}) and (\ref{lst:pipeline_bash}) a pipeline is used for MFCC and pitch extraction on three utterances from two speakers, the two scripts in Python and bash being strictly equivalent and giving the same result. For each utterance, the extracted features have 16 dimensions: 13 for MFCC and 3 for pitch estimates.}
\end{figure*}

\subsection{High-level extraction pipeline}
\label{sec:pipeline}

The modular design described above allows the creation of arbitrary pipelines involving multiple steps, such as raw features extraction, pitch estimation and normalization. In order to simplify the use of such complex pipelines, Shennong exposes a high-level interface made of three steps, which can be used from Python using the \texttt{pipeline} module or from command line using the \texttt{speech-features} program.

The first step is to define a list of utterances on which to apply the pipeline, as a list of audio files, with optional utterances name, speaker identification and onset/offset times. The second step is to configure the extraction pipeline by selecting the extraction algorithms to use. This step generates a configuration with default parameters, which can further be edited by the user. The third and last step is to apply the configured pipeline on the defined utterances.
Two use cases are illustrated in Fig. \ref{fig:use_example}: the use of the low-level API to extract MFCCs on a single file (Fig. \ref{lst:mfcc_simple}) and the use of a high-level pipeline to extract both MFCCs and pitch on three utterances from two speakers, from the Python API (Fig. \ref{lst:pipeline_python}) and command line (Fig. \ref{lst:pipeline_bash}).

\section{Applications}
\label{sec:experimentation}

This section demonstrates the use of Shennong for experimental purpose. Three applications are introduced: a comparison a features extraction algorithms on a phones discrimination task, an analysis of the VTLN model performance as a function of speech duration used for training and a comparison of pitch estimation algorithms on various noise conditions. The code to replicate those experiments is distributed with Shennong\footnote{\url{https://github.com/bootphon/shennong/tree/v1.0/examples}}.

\subsection{Phones discrimination task}
\label{sec:features}

\begin{table*}[ht]
\begin{subfigure}{\textwidth}
\centering
\resizebox{\textwidth}{!}{%
\begin{tabular}{|l||c|c|c||c|c|c||c|c|c||c|c|c|}
    \hline
    \multirow{3}{*}{Algorithm} &
    \multicolumn{6}{|c||}{Within speakers} &
    \multicolumn{6}{|c|}{Across speakers} \\
    \cline{2-13} &
    \multicolumn{3}{|c||}{without VTLN} &
    \multicolumn{3}{|c||}{with VTLN} &
    \multicolumn{3}{|c||}{without VTLN} &
    \multicolumn{3}{|c|}{with VTLN} \\
    \cline{2-13} &
    raw & $+\Delta$/F0 & +CMNV &
    raw & $+\Delta$/F0 & +CMNV &
    raw & $+\Delta$/F0 & +CMNV &
    raw & $+\Delta$/F0 & +CMNV \\ \hline
    Spectrogram & 16.7 & 15.2 & 20.2 & - & - & - & 30.3 & 27.9 & 29.7 & - & - & - \\ \hline
    Filterbank & 12.8 & \textbf{11.6} & 18.2 & 12.6 & \textbf{11.4} & 18.1 & 24.9 & \textbf{22.1} & 26.5 & 23.2 & 20.7 & 25.4 \\ \hline
    MFCC & 13.0 & 12.5 & 12.4 & 12.8 & 12.3 & 12.0 & 27.2 & 26.4 & 24.0 & 23.4 & 22.7 & 20.0 \\ \hline
    PLP & 12.5 & 12.4 & 11.9 & 12.5 & 12.4 & 11.9 & 28.0 & 26.6 & 23.8 & 24.7 & 23.5 & \textbf{19.7} \\ \hline
    Rasta-PLP & 14.3 & 14.2 & 12.5 & 14.2 & 14.1 & 12.5 & 28.5 & 26.8 & 25.3 & 24.6 & 23.6 & 21.3 \\ \hline
    Bottleneck & \textbf{8.5} & 8.5 & 8.6 & - & - & - & \textbf{12.5} & 12.5 & 12.5 & - & - & - \\ \hline
\end{tabular}}
\caption{ABX scores for English}
\end{subfigure}
\begin{subfigure}{\textwidth}
\centering
\resizebox{\textwidth}{!}{%
\begin{tabular}{|l||c|c|c||c|c|c||c|c|c||c|c|c|}
    \hline
    \multirow{3}{*}{Algorithm} &
    \multicolumn{6}{|c||}{Within speakers} &
    \multicolumn{6}{|c|}{Across speakers} \\
    \cline{2-13} &
    \multicolumn{3}{|c||}{without VTLN} &
    \multicolumn{3}{|c||}{with VTLN} &
    \multicolumn{3}{|c||}{without VTLN} &
    \multicolumn{3}{|c|}{with VTLN} \\
    \cline{2-13} &
    raw & $+\Delta$/F0 & +CMNV &
    raw & $+\Delta$/F0 & +CMNV &
    raw & $+\Delta$/F0 & +CMNV &
    raw & $+\Delta$/F0 & +CMNV \\ \hline
    Spectrogram & 19.2 & 16.8 & 19.2 & - & - & - & 34.6 & 32.0 & 26.5 & - & - & - \\ \hline
    Filterbank & 13.8 & \textbf{11.7} & 15.2 & 13.6 & \textbf{11.4} & 15.2 & 28.1 & 25.1 & \textbf{21.5} & 26.9 & 24.0 & \textbf{20.7} \\ \hline
    MFCC & 17.1 & 16.2 & 14.6 & 17.5 & 16.5 & 14.6 & 33.6 & 32.8 & 26.0 & 31.4 & 30.6 & 22.7 \\ \hline
    PLP & 16.2 & 14.6 & 14.0 & 16.2 & 14.7 & 14.2 & 33.5 & 31.2 & 26.2 & 31.7 & 29.5 & 22.2 \\ \hline
    Rasta-PLP & 13.7 & 12.5 & 12.3 & 13.5 & 12.2 & 12.0 & 27.9 & 25.2 & 23.9 & 25.0 & 22.8 & 21.7 \\ \hline
    Bottleneck & \textbf{6.9} & 7.0 & 7.3 & - & - & - & \textbf{9.5} & 9.6 & 9.6 & - & - & - \\ \hline
\end{tabular}}
\caption{ABX scores for Xitsonga}
\end{subfigure}
\caption{\label{table:abx}Comparison of features extraction algorithms on a phones discrimination task, within and across speakers, with and without VTLN, for English and Xitsonga datasets. Scores are ABX error rates in \% (random score is 50\%). The \emph{raw} configuration is based on raw features alone. The \emph{+$\Delta$/F0} adds first/second order derivatives and Kaldi pitch estimates. The \emph{+CMVN} adds a CMVN normalization by speaker on top of \emph{+$\Delta$/F0}. VTLN is not available for spectrogram and bottleneck features. Best scores for each configuration are in bold font.}
\end{table*}

This section details a phones discrimination experiment used as a proxy to compare the features algorithms available in Shennong. It reproduces the track 1 of the Zero Speech Challenge 2015 \cite{zerospeech2015} using the same datasets and evaluation.

\subsubsection{Methods}

The datasets are composed of selected segments from two free, open access and annotated speech corpora, the Buckeye Corpus \cite{buckeye} (American English, 12 speakers, 10h34m44s) and the NCHLT Speech Corpus \cite{xitsonga} (Xitsonga, 24 speakers, 4h24h37s). The gold phonemes transcriptions have been obtained from a forced alignment using Kaldi.

The evaluation of phone discriminability uses a minimal pair ABX task, a psychophysically inspired algorithm that only requires a notion of distance between the representations of speech segments \cite{schatz2019early,schatz2013,schatz2014}.
The ABX discriminalbility, for example, between $[apa]$ and $[aba]$, is defined as the probability that the representations of $A$ and $X$ are more similar than representations of $B$ and $X$, over all triplets of tokens such that $A$ and $X$ are tokens of $[aba]$ and $B$ a token of $[apa]$. The discriminability is evaluated \emph{within} speakers, where $A$, $B$ and $X$ are uttered by the same speaker, and \emph{across} speakers, such that $X$ is uttered by a different speaker than $A$ and $B$.
The global ABX phone discriminability score aggregates over the entire set of minimal triphone pairs such as $([aba], [apa])$ to be found in the dataset. The metric used for ABX evaluation is the Dynamic Time Wrapping divergence using the cosine distance as underlying frame-level metric.

The following features extraction algorithms are considered: spectrogram, filterbank, MFCC, PLP, RASTA-PLP and multilingual bottleneck network. All the algorithms are used with default arguments. Each algorithm is declined in three pipeline configurations. The raw features alone are first considered, noted as \emph{raw} in Table \ref{table:abx}, and of dimension $n$. Then the concatenation of the raw features with their first and second order derivatives, along with pitch estimates, are used and noted $+\Delta/F0$, giving a dimension $3n+3$. The cross-correlation pitch estimation algorithm from Kaldi is used, it outputs three channels: probability of voicing, normalized log pitch and raw pitch derivative. Finally CMVN is applied on a per-speaker basis on the $+\Delta/F0$ configuration, giving a zero mean and unitary variance on each channel independently, and is noted as $+CMVN$. Furthermore a VTLN model is trained on 10 minutes of speech per speaker for each of the two corpora and is applied to spectrogram, filterbank, MFCC and PLP, for each of the three pipeline configurations.

\subsubsection{Results}

Experimental results are presented in Table \ref{table:abx}. First considering the overall scores, the bottleneck neural network largely outperforms the spectro-temporal algorithms in every configuration. This is expected as the bottleneck model is trained for phone discrimination. Among the spectro-temporal algorithms, the filterbank model performs very well and reachs the best score on 7 over 8 configurations. This result is to be underlined as it beats MFCC, which is by far the most used algorithm in the literature, in all the configurations excepted on English across speakers with VTLN.

Now considering the impact of $raw$, $+\Delta/F0$ and $+CMNV$ pipelines for the different algorithms, it is demonstrated that adding pitch, deltas and CMVN to raw features is beneficial for both MFCC, PLP and Rasta-PLP is all configurations, excepted for the bottleneck algorithm. Spectrogram and filterbank algorithms benefit from pitch and deltas as well but, with the exception of the Xitsonga across speakers configuration, the addition of CMVN degrades the ABX score. Rasta filtering on PLP gives different results across languages: it degrades the score on English but improve them on Xitsonga.

Finally considering the use of VTLN, it improves both MFCC and PLP scores by about 4\% on the across speakers context, whereas filterbank gain about 1\%. No or little improvement is attested within speakers for all the algorithms. This is expected in this context because ABX scores are computed on a single speaker.

\subsection{VTLN model training}
\label{sec:vtln}

This section explores the influence of the amount of speech duration used for VTLN training on the resulting VTLN coefficients and phones discriminability scores.

\subsubsection{Methods}

The same segment of the Buckeye English corpus as in Section \ref{sec:features} is used. It is composed of 10h34m44s of speech balanced across 12 speakers. In order to train several VTLN models on variable speech duration, this corpus is split in sub-corpora containing a given speech duration per speaker. The considered durations are 5s, 10s, 20s, 30s, 60s and up to 600s by steps of 60s. The subcorpora are built without overlap: the first block of fixed duration for each speaker are joined together, then for the second blocks, etc. This gives a total of 1010 corpora, from 479 for 5s per speaker to 2 for 600s per speaker, following a power law. For each of those corpora, a VTLN model is trained using the default parameters and VTLN coefficients are extracted.

Then MFCC features are extracted from those corpora, using default parameters, and normalized with their associated VTLN coefficients. MFCC features are declined over the 3 pipeline configurations $raw$, $+\Delta/F0$ and $+CMVN$, as detailed in Section \ref{sec:features}. ABX discriminability score is then computed across speakers as before. To mitigate the computational cost, a maximum of 10 corpora per duration are randomly sampled and considered for both MFCC extraction and ABX scoring. Moreover MFCC and ABX are computed without VTLN, and with a VTLN model trained on the entire corpus, giving a total of 102 discriminability scores for all the considered durations.

\subsubsection{Results}

\begin{figure}[!t]
    \centering
    \includegraphics[width=\linewidth]{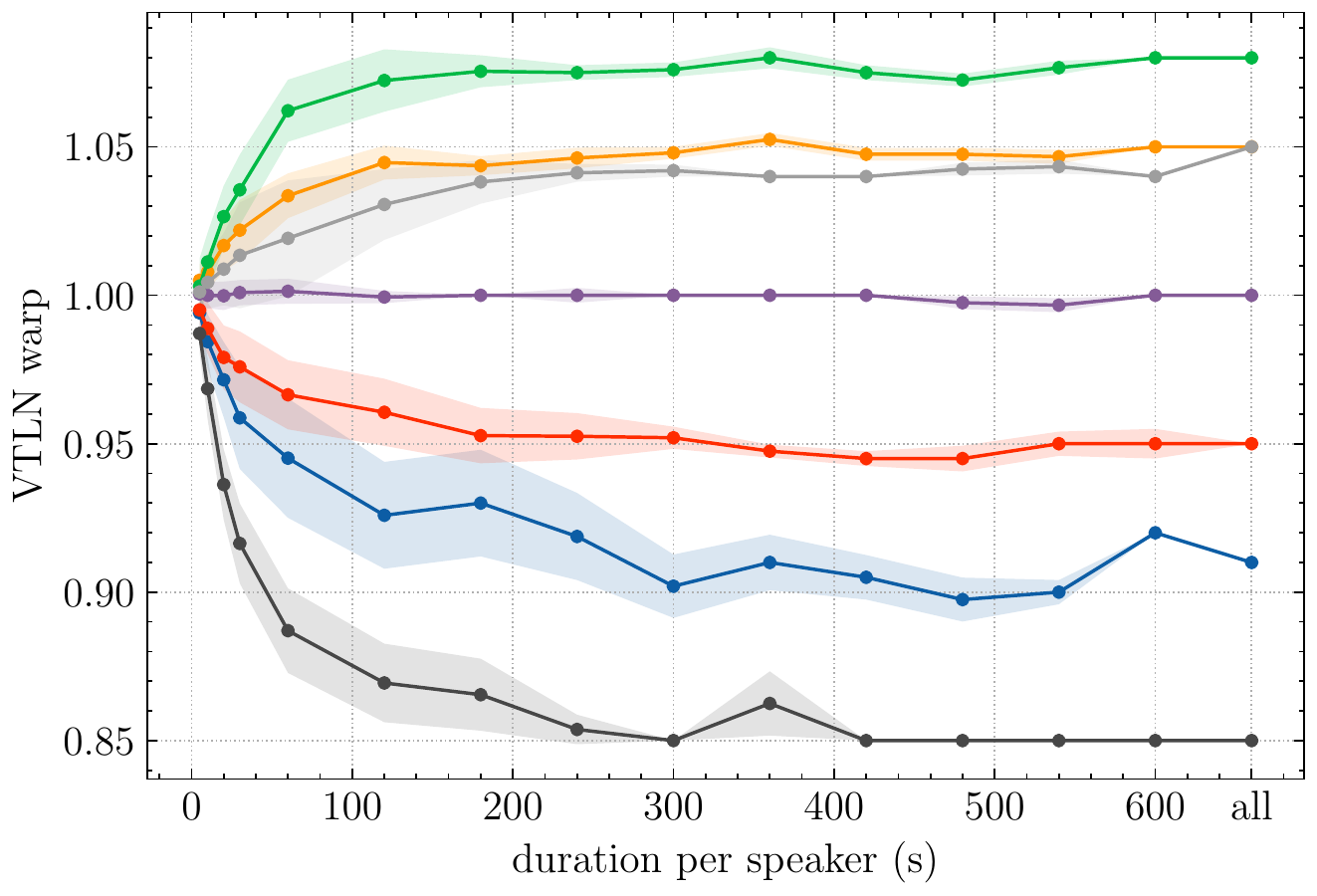}
    \caption{Average VTLN coefficients for different speakers according to the speech duration per speaker used for VTLN training. The lines correspond to coefficients of 7 representative speakers out of 12. The shaded areas are standard deviations.}
    \label{fig:vtln_warps}
\end{figure}

\begin{figure}[!t]
    \centering
    \includegraphics[width=\linewidth]{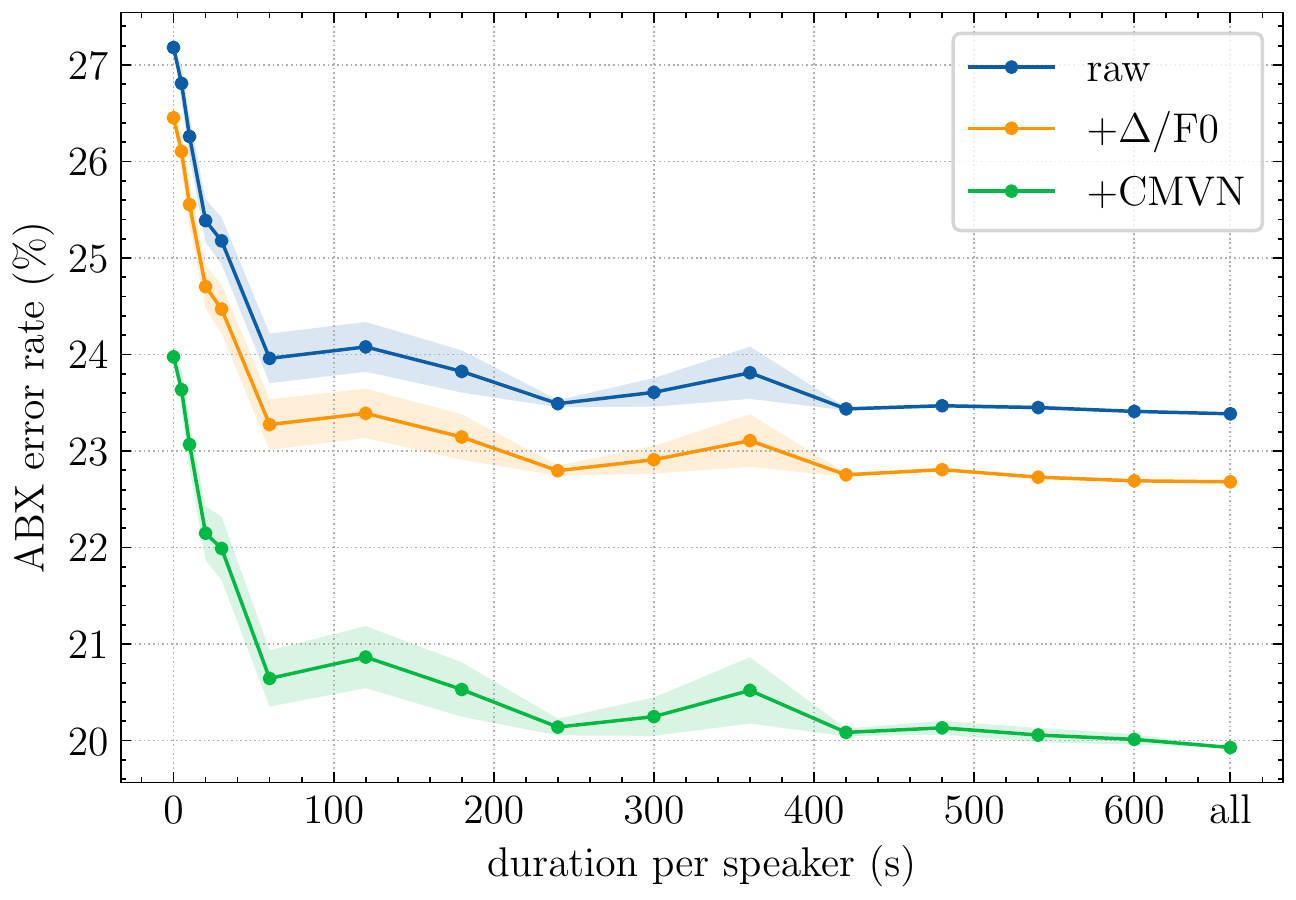}
    \caption{ABX error rate obtained for MFCC features across speakers on English, for 3 pipeline configurations, and various speech duration per speaker used for VTLN training. The shaded areas are standard deviation.}
    \label{fig:vtln_abx}
\end{figure}

Figure \ref{fig:vtln_warps} shows the evolution of the VTLN coefficients for different speakers, in function of the amount of speech per speaker used for training. With 300s per speaker, or 1h of speech in total, the coefficient are largely converged and remain overall stable when more data is added for training. This demonstrates that training a VTLN does not require a large amount of data, thus reducing the computational needs and training time. Moreover the differentiation comes very early: with 30s of speech per speaker only, the VTLN coefficients are already clustered.

Figure \ref{fig:vtln_abx} shows the ABX error rate obtained on MFCC features without VTLN normalization and with VTLN computed using different speech durations per speaker. First considering the scores obtained without VTLN and with VTLN trained on the whole dataset, results match those displayed in Table \ref{table:abx}: in $raw$ configuration the scores go from 27.2\% to 23.4\%, from 26.4\% to 22.7\% for $+\Delta/F0$ and from 24.0\% to 20.0\% for $+CMVN$. The 3 configurations follow the same tendency and rapidly converge to a nearly optimal score, starting with 60s of speech per speaker for VTLN training. Consolidating from results on Figure \ref{fig:vtln_warps}, it is shown here that the VTLN coefficients do not need to have fully converged to yield a close to optimal normalization.

\subsection{Pitch estimation}
\label{sec:pitch}

\begin{figure*}[ht]
\begin{subfigure}{0.49\textwidth}
    \includegraphics[width=\linewidth]{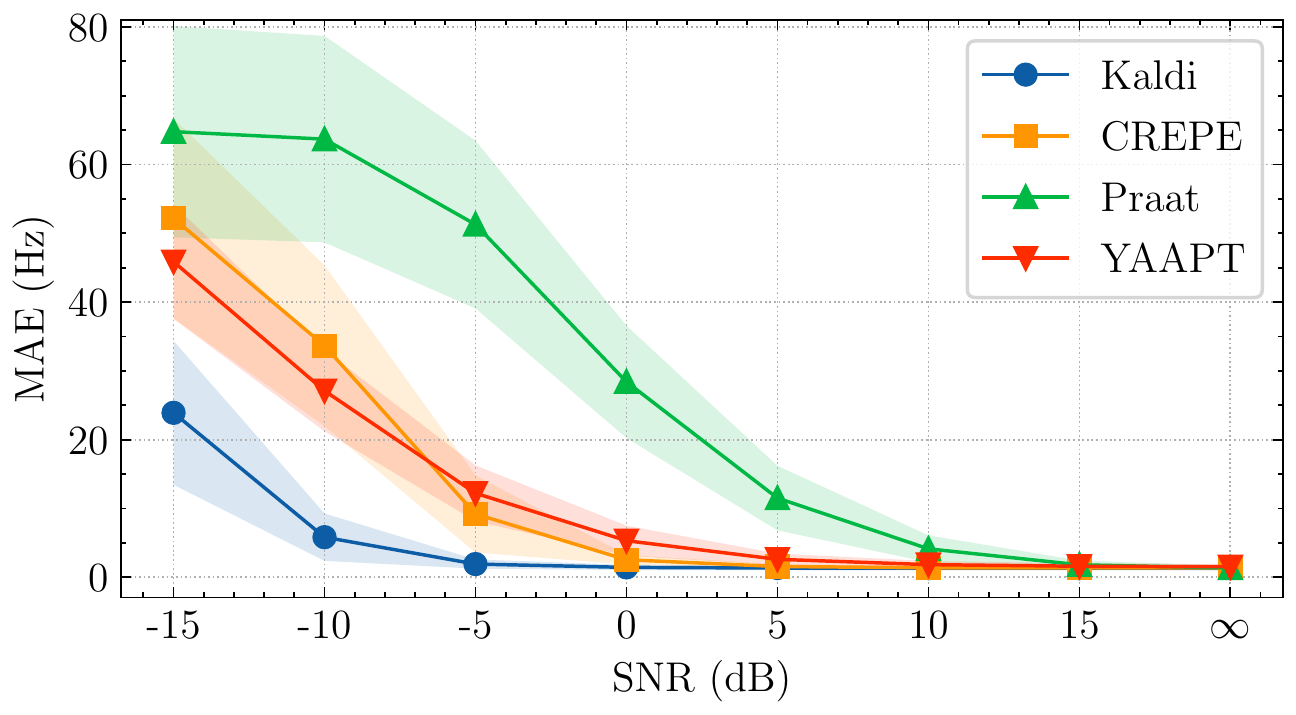}
    \caption{Gaussian noise, Mean Absolute Error}
    \label{fig:pitch:maegauss}
\end{subfigure}
\begin{subfigure}{0.49\textwidth}
    \includegraphics[width=\linewidth]{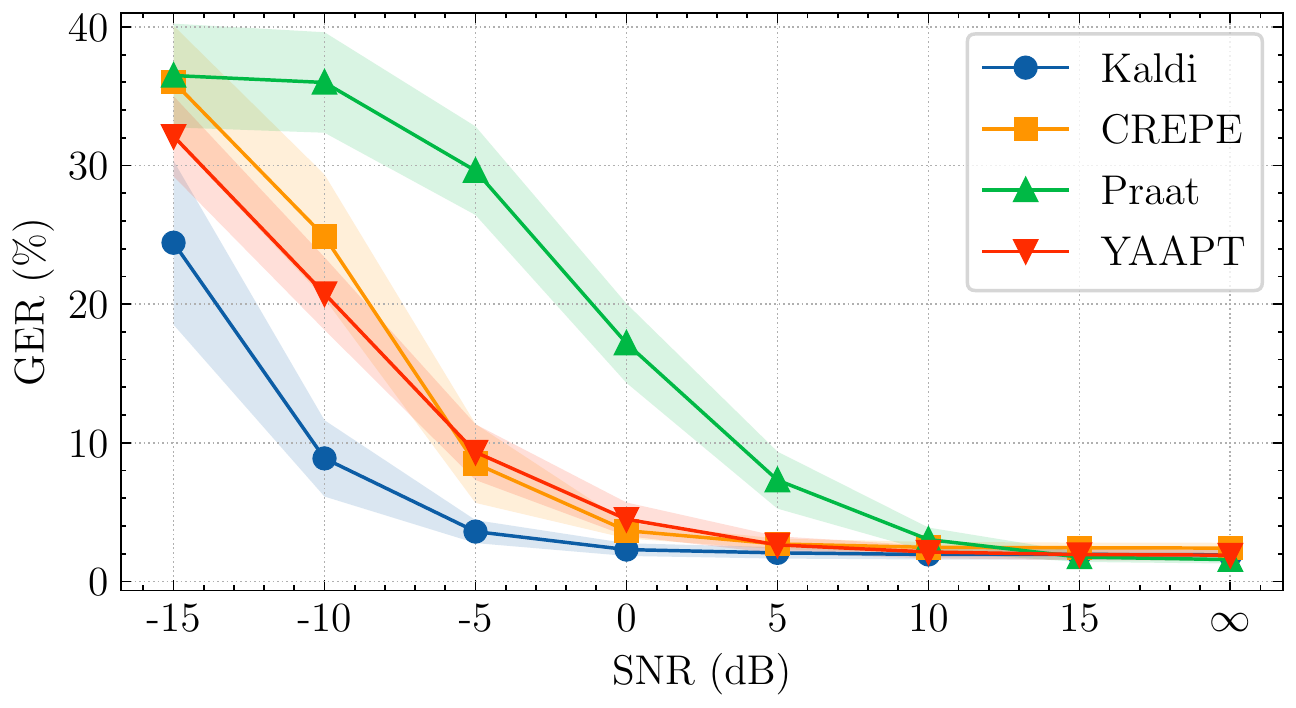}
    \caption{Gaussian noise, Gross Error Rate}
    \label{fig:pitch:gergauss}
\end{subfigure}
\begin{subfigure}{0.49\textwidth}
    \includegraphics[width=\linewidth]{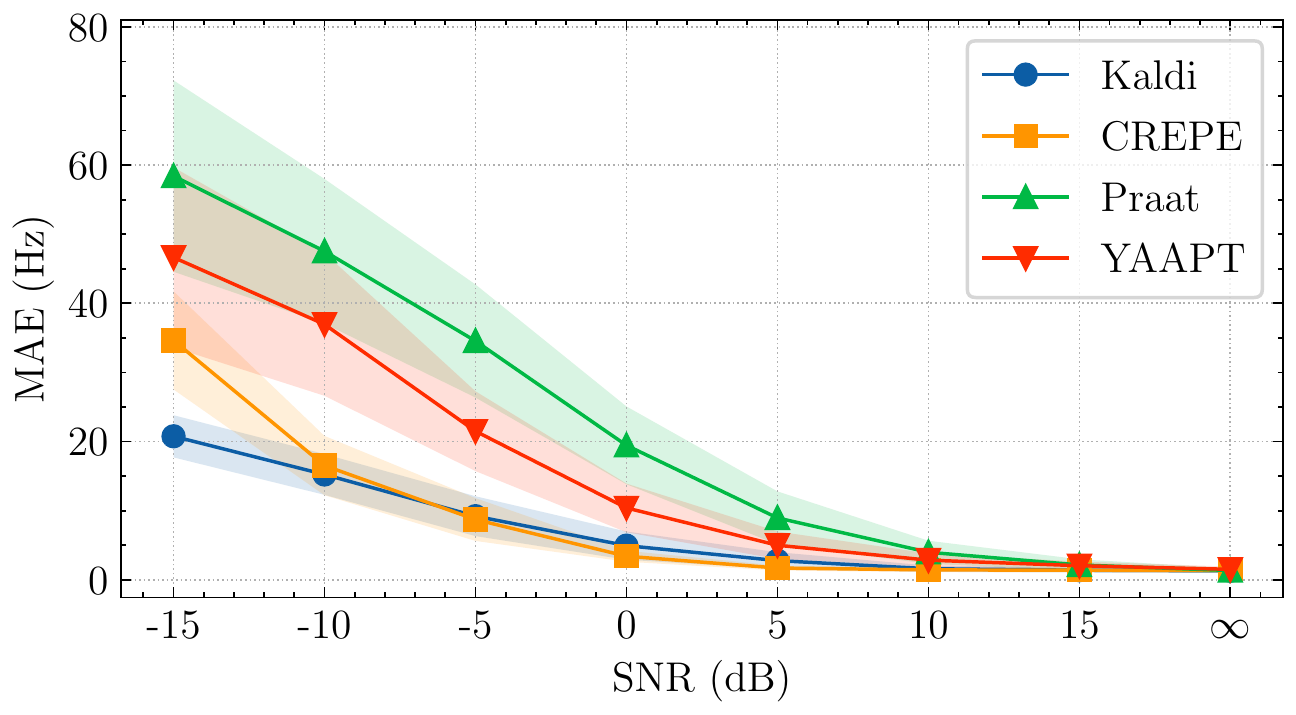}
    \caption{Babble noise, Mean Absolute Error}
    \label{fig:pitch:maebabble}
\end{subfigure}
\begin{subfigure}{0.49\textwidth}
    \includegraphics[width=\linewidth]{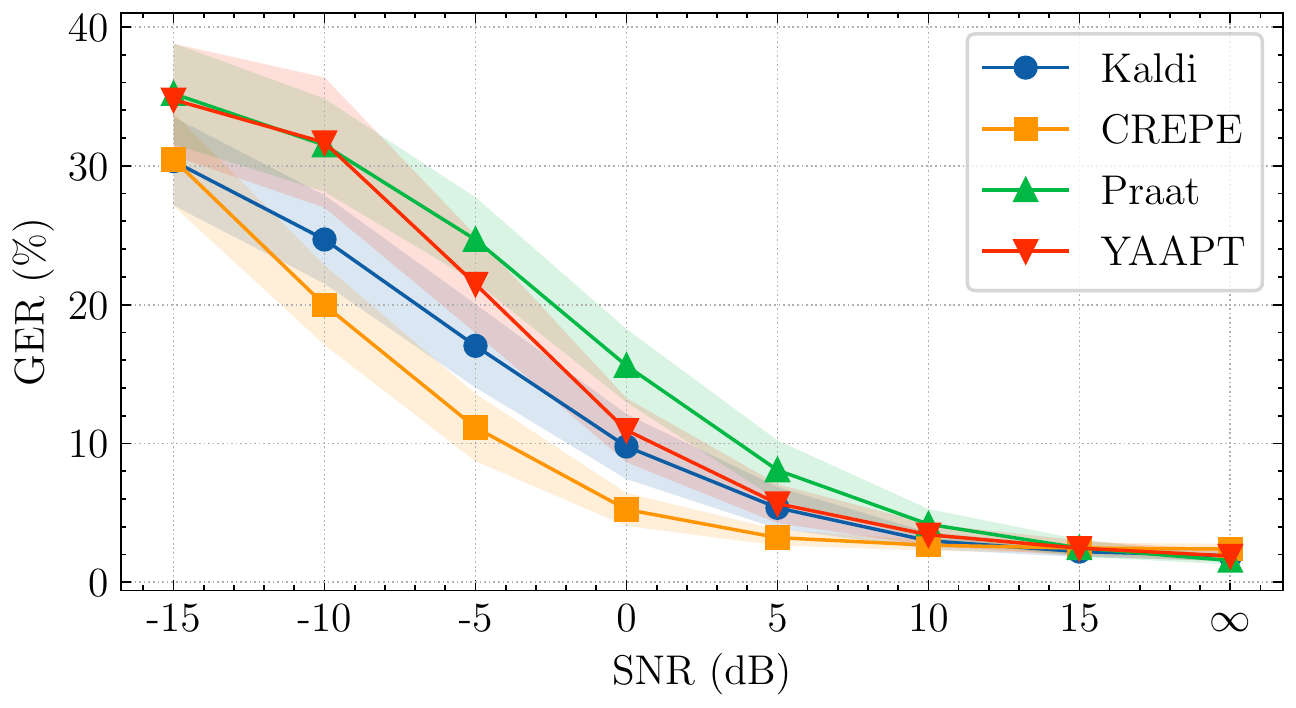}
    \caption{Babble noise, Gross Error Rate}
    \label{fig:pitch:gerbabble}
\end{subfigure}
\caption{\label{fig:pitch}Pitch estimation error for the Kaldi, CREPE, Praat and YAAPT algorithms using Mean Absolute Error and Gross Error Rate on the KEELE dataset. White Gaussian noise or babble noise is added at various signal to noise ratios. Lines are mean errors over all the 10 speakers and shaded areas are standard deviations.}
\label{fig:exp_pitch}
\end{figure*}

This section compares pitch estimation algorithms on speech, under various noise conditions.

\subsubsection{Methods}

The Keele Pitch Database \cite{keele} is used for evaluation. It consists of approximately 6 minutes of clean speech with pitch estimates, separated in 10 phonetically balanced sentences by five male and five female speakers. The pitch is estimated from the autocorrelation of a laryngograph signal using frames of 25.6~ms with a 10~ms overlap. As noise robustness is key to applications with real world data, the KEELE dataset has been corrupted by additive noise at 7 signal to noise ratios (SNR) ranging from -15~dB to 15~dB. Two noises are considered: white Gaussian noise and babble noise, which consist in a recording of a restaurant ambiance.

The Kaldi and CREPE pitch estimators from Shennong (see Section \ref{sec:models}) are compared with two other popular models: the Praat algorithm \cite{praatpitch,Praat}, which uses an auto-correlation method, and the YAAPT algorithm \cite{yaapt} based on a combination of time and frequency domain processing using normalized cross-correlation. To match the gold pitch estimates from the KEELE dataset, the Kaldi, CREPE and YAAPT algorithms are parametrized to use frames of 25.6ms with a 10ms overlap. The Praat algorithm does not support frames parametrization so its estimates have been linearly interpolated to match the gold timestamps.

Finally an important amount of frames is estimated as unvoiced on clean speech: only 50.3~\% of the KEELE gold estimates are valid pitches. Other values correspond to an absence of voiced speech or a corrupted laryngograph signal. The algorithms as well estimate some frames as unvoiced. This is detected by a pitch estimate at 0 for Praat and YAAPT models, or by a low confidence for Kaldi and CREPE. To avoid estimation biases, the union of all the frames classified as unvoiced within the KEELE dataset and by the 4 algorithms on clean speech are removed from the evaluation. This leads to 36.3~\% of the total number of frames being conserved for the evaluations at different SNR.
The behavior of the four studied algorithms differs for frames classified as unvoiced: Kaldi and CREPE always give a positive pitch estimate but associated to a low confidence, whereas Praat and YAAPT estimates a pitch at 0. As such, when frames are estimated unvoiced on noisy conditions for Praat and YAAPT, they are evaluated against non-zero ground truth.


Two performance measures are considered. The Gross Error Ratio (GER) is the proportion of pitch estimates that differ from more than 5~\% from the ground truth. The Mean Absolute Error (MAE) is the mean of the absolute error between the pitch estimates and the ground truth. Thus, given a speech signal with $n$ frames, $x\in\mathds{R}^n$ its ground truth and $\tilde{x}\in\mathds{R}^n$ its pitch estimates for each frame, the GER and MAE metrics are expressed as follows:
\begin{equation}
    MAE(\tilde{x}, x) = \frac{1}{n}\sum_{i=1}^n|\tilde{x}_i-x_i|,
\end{equation}\begin{equation}
    GER(\tilde{x}, x) = \frac{100}{n}\sum_{i=1}^n\mathds{1}_{|\tilde{x}_i - x_i|>0.05x_i},
\end{equation}
where $\mathds{1}_{p(.)}$ is 1 when the predicate $p(.)$ is true and 0 otherwise.

\subsubsection{Results}

Figure \ref{fig:pitch} shows the evaluation error obtained for the four algorithms and the two noises at the considered SNR, for both MAE and GER metrics.
Considering first the errors obtained on Gaussian noise (Figures \ref{fig:pitch:maegauss} and \ref{fig:pitch:gergauss}), MAE and GER follow similar patterns for all the algorithms. When there is no or little noise, all models obtain a low error which is stable across speakers. The Praat algorithm is the first one to have degraded performances, starting at 10~dB, where CREPE and YAAPT starts at 5~dB. The Kaldi algorithms is particularly strong against this noise, with a stable error up to -5~dB. CREPE and YAAPT have similar errors across SNR, with CREPE being more robust to noise up to -5~dB and YAAPT below -5~dB. When considering the effect of additive babble noise (Figures \ref{fig:pitch:maebabble} and \ref{fig:pitch:gerbabble}), the algorithms performance starts to decrease at 10~dB. Across the SNR range, Kaldi and CREPE performances are very close and give the lowest errors. CREPE is more reliable with a lowest GER, excepted at -15~dB where Kaldi performs better. YAAPT and primarily Praat do not perform well on babbbke noise and have high errors rates and standard deviations.

Overall, Kaldi, CREPE and YAAPT perform better on Gaussian noise than babble noise, YAAPT being the most sensitive to the latter. Praat is the only algorithm to have an increase of performance on babble noise. Finally, both CREPE and Kaldi appears to being more reliable estimators than Praat and YAAPT, with Kaldi being more robust to Gaussian noise and CREPE to babble noise.

\section{Discussion}
\label{sec:discussion}

This paper introduced Shennong, an open source Python package for speech features extraction. The toolbox covers a wide range of well-established state of the art algorithms, most of them being implemented after Kaldi. The software architecture and components of Shennong focuses on ease-of-use, reliability and extensibility. It covers different use cases: few commands are sufficient to configure and apply a complex extraction pipeline, but power-users can benefit from the Python API to hand-tune any part of the pipeline or to integrate Shennong in their own projects.

Three experiments on features comparison using Shennong are provided, showing that the package is mature enough to be used as a research tool. The first experiment demonstrates that, counter-intuitively, mel filterbanks perform better that the popular MFCC on a phones discrimination task. It also showed that VTLN speaker normalization reduces the error rates up to 5~\%. The second experiment analyzed the amount of speech required to train a VTLN model and demonstrates than 5 to 10 minutes of signal per speaker are enough to reach near-optimal performances. Finally the last experiment compared pitch estimation algorithms under various noise conditions and demonstrated than the two algorithms provided with Shennong are more robust to noise than YAAPT and Praat algorithms, commonly used in phonology.

The development of Shennong is not over as more features extraction algorithms are planned to be added, such as Voice Activity Detection and Contrastive Predictive Coding \cite{cpc}. Because Shennong is free software with open sources, it's future will be impacted by the users needs and requests. We hope the Shennong's community of users and contributors will grow as its visibility increase.

\begin{acknowledgements}
This work is founded by Inria (Grant ADT-193), the Agence Nationale de la Recherche (ANR-17-EURE-0017 Frontcog, ANR-10-IDEX-0001-02 PSL, ANR-19-P3IA-0001 PRAIRIE 3IA Institute), CIFAR (Learning in Minds and Brains) and Facebook AI Research (Research Grant).
\end{acknowledgements}

\section*{Conflict of interest}
The authors declare that they have no conflict of interest.

\bibliographystyle{spmpsci}      
\bibliography{biblio}

\end{document}